\documentclass[conference]{IEEEtran}
\IEEEoverridecommandlockouts
\usepackage{cite}
\usepackage{amsmath,amssymb,amsfonts}
\usepackage{algorithmic}
\usepackage{graphicx}
\usepackage{textcomp}
\usepackage{xcolor}
    
\usepackage{booktabs}
\usepackage{bbding}
\usepackage{multirow}
\usepackage{hyperref}
\hypersetup{
    colorlinks=true,
    linkcolor=black,
    filecolor=magenta,      
    urlcolor=black,
    citecolor = black,
    pdfpagemode=FullScreen,
    }
\input{alphabet}
\definecolor{mypurple}{rgb}{0.8,0.5,0.8}
\definecolor{myviolet}{rgb}{0.6,0.3,0.4}
\definecolor{cadmiumgreen}{rgb}{0.0, 0.42, 0.24}
\newcommand{\YZ}[1]{\textcolor{black}{#1}}

\begin{document}
\bstctlcite{IEEEexample:BSTcontrol}  
\title{
Ultrasound Image Enhancement with the Variance of Diffusion Models\\
\thanks{This work has been supported by the European Regional Development Fund (FEDER), the Pays de la Loire Region on the Connect Talent scheme (MILCOM Project) and Nantes Métropole (Convention 2017-10470).}
}
\author{
\IEEEauthorblockN{Yuxin Zhang\IEEEauthorrefmark{1}, Clément Huneau\IEEEauthorrefmark{1}, Jérôme Idier\IEEEauthorrefmark{1} and Diana Mateus\IEEEauthorrefmark{1}}
\IEEEauthorblockA{\IEEEauthorrefmark{1}Nantes Université, École Centrale Nantes, LS2N,
CNRS, UMR 6004, F-44000 Nantes, France \\
Email: yuxin.zhang@ls2n.fr}
}
\maketitle

\begin{abstract}
Ultrasound imaging, despite its widespread use in medicine, often suffers from various sources of noise and artifacts that impact the signal-to-noise ratio and overall image quality. Enhancing ultrasound images requires a delicate balance between contrast, resolution, and speckle preservation. This paper introduces a novel approach that integrates adaptive beamforming with denoising diffusion-based variance imaging to address this challenge. \YZ{By applying Eigenspace-Based Minimum Variance (EBMV) beamforming and employing a denoising diffusion model fine-tuned on ultrasound data}, our method computes the variance across multiple diffusion-denoised samples to produce high-quality despeckled images. This approach leverages both the inherent multiplicative noise of ultrasound and the stochastic nature of diffusion models. Experimental results on a publicly available dataset demonstrate the effectiveness of our method in achieving superior image reconstructions from single plane-wave acquisitions. The code is available at: \href{https://github.com/Yuxin-Zhang-Jasmine/IUS2024_Diffusion}{https://github.com/Yuxin-Zhang-Jasmine/IUS2024\_Diffusion}.
\end{abstract}


\section{Introduction}
Ultrasound (US) imaging is a widely used diagnostic tool due to its real-time capabilities, affordability, portability, and non-ionizing nature. These advantages make it a preferred option over Magnetic Resonance (MR) and Computed Tomography (CT) in fields such as cardiology and obstetrics. Despite these benefits, US imaging is often challenged by electronic noise, acoustic attenuation, and artifacts like reverberation, shadowing, and speckle. 

The standard Delay-and-Sum (DAS)~\cite{DAS} beamforming algorithm converts channel signals into B-mode images, prioritizing speed over image quality. To enhance US image quality, various advanced techniques have been developed, including adaptive beamforming methods~\cite{MV, EBMV, iMAP}, model-based approaches~\cite{IPB_Ozkan, RED_USIPB, NDT_laroche}, and (physics-informed) deep learning (DL) techniques~\cite{ABLE, Jingke2021, CNNperdios2022, DL_van_sloun_2020, DGM4MICCAI}. Although these methods can produce high-quality reconstructions, they often regard US speckle as useful information since speckle can assist in motion tracking by indicating tissue movement. However, in applications such as organ and tumor segmentation or classification, speckle can be problematic~\cite{DeclutterCNN2018}.

Speckle noise in US imaging is caused by the interference of coherent US waves scattered by small tissue structures, creating a granular pattern. Various despeckling techniques have been developed, 
such as Anisotropic Diffusion (AD)-based methods~\cite{AD2006, AD2007, ADMSS}, wavelet-based techniques~\cite{Wavelet2006, Wavelet2007, Wavelet2013}, the Bilateral filter~\cite{BF2010},  Nonlocal Means~\cite{NLmeans2009}, and more recently, DL-based approaches~\cite{DeclutterCNN2018, DeclutterCNN2020}.
However, these methods often overlook the presence of electronic noise, which can be significant, especially for ultrafast unfocused emissions such as Plane Waves (PWs). Additionally, they typically operate on processed US images, such as enveloped absolute or log-compressed data, rather than the original signed signals, limiting the preservation of signal characteristics.

\YZ{Stochastic} diffusion models, renowned for image synthesis~\cite{BGAN}, have recently demonstrated their strong capability in US imaging~\cite{DGM4MICCAI,dehaze,DenoDDPM1,DenoDDPM2,DiffusionIPB_HLan}.  
\YZ{In our previous work~\cite{EUSIPCO}, we have demonstrated that computing the variance across multiple diffusion-reconstructed samples effectively achieves despeckling without over-smoothing. Specifically, US image reconstruction was framed as an inverse problem with the DAS-beamformed Radio-Frequency (RF) image serving as the measurement. Using the Denoising Diffusion Restoration Models (DDRM)~\cite{DDRM} framework, each run estimated the tissue reflectivity, and the variance of multiple independent estimations was used to produce the final despeckled image.} However, the high complexity of the model matrix slowed diffusion sampling and made the Singular Value Decomposition (SVD) \YZ{required by DDRM} both memory- and time-intensive.

\YZ{In the present work, we use a denoising model rather than an inverse problem model and replace the DAS-beamformed image with an adaptively beamformed image. By applying diffusion denoising multiple times and calculating the variance across these denoised samples, we can produce a high-resolution despeckled image more efficiently, addressing the complexity and computational challenges observed in our previous work.}
Our contributions are: 
1) Introducing an adaptive beamforming-based diffusion variance imaging, which achieves faster sampling and competitive image quality. 
2) Showing the complementary effects of combining pixel-wise beamforming with denoising diffusion variance imaging, particularly for resolution improvement and background recovery.

\section{Method}
\begin{figure}[tb]
    \centering
    \includegraphics[width=\linewidth]{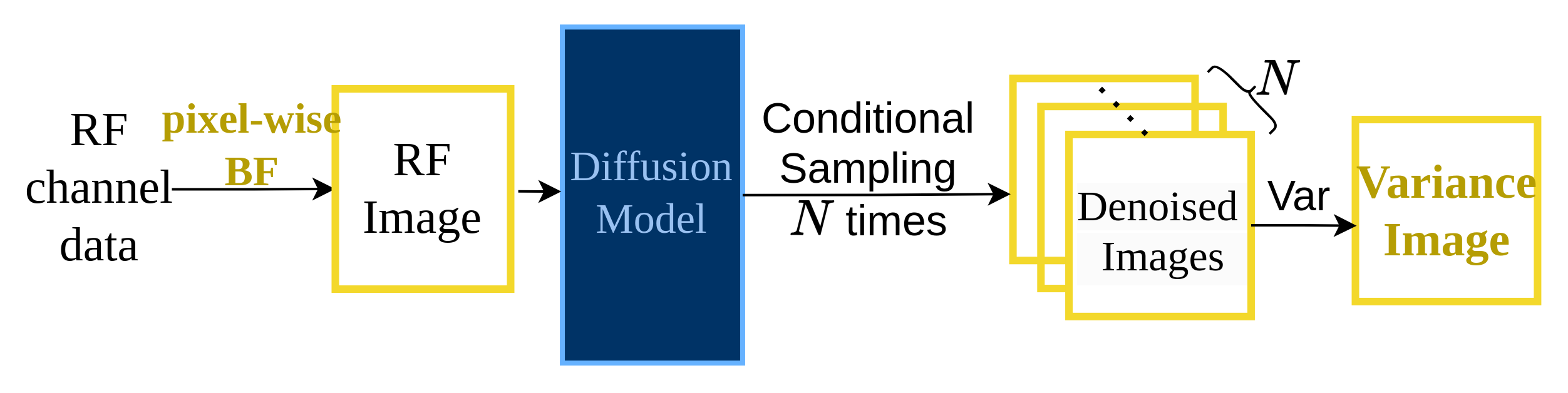}
    \caption{Method overview}
    \label{fig: overview}
\end{figure}

\YZ{The proposed method} aims to remove both electronic and speckle noise from \YZ{the received signals after a single PW transmission}, see Fig.~\ref{fig: overview} for an overview. Initially, an adaptive pixel-wise beamformer is applied to convert the acquired time-domain RF channel data into the spatial domain. \YZ{Subsequently,} by running the conditional diffusion generative process multiple times and calculating the variance of the generated samples, an enhanced image is produced.

\subsection{Adaptive Beamforming}
\YZ{Convolutions between the excitation pulse, transducer impulse responses (both transmit and receive), and the spatial impulse responses result in a spatially variant point-spread effect in the DAS-beamformed image. This convolutional effect is carefully modeled in \cite{EUSIPCO} to achieve highly accurate reconstruction, albeit at the cost of increased computational complexity. In contrast, the present work mitigates this degradation by using the adaptively beamformed RF image as the measurement, simplifying the diffusion sampling process from solving an inverse problem to straightforward denoising. This approach enables faster diffusion sampling while maintaining image quality.} Among the various pixel-wise adaptive beamforming approaches~\cite{MV, EBMV, iMAP}, this work employs the Eigenspace-Based Minimum Variance (EBMV)~\cite{EBMV} technique.

The EBMV beamformer builds upon the \YZ{Minimum Variance (MV)} beamformer~\cite{MV}. Let the delayed RF channel data \YZ{for one pixel} be denoted as $\yv \in \mathbb{R}^{{\it Ne} \times Np}$, where ${\it Ne}$ and $Np$ represent the number of channels and the number of time samples, respectively. The channel weights for each pixel in MV beamforming are calculated as follows:
\begin{equation}
    \wv_{\it MV} = \frac{\Rv^{-1}\mathbf{1}}{\mathbf{1}^{\mathrm{H}}\Rv^{-1}\mathbf{1}},
\end{equation}
where $\wv_{\it MV} \in \mathbb{R}^{{\it Ne} \times 1}$ and $\Rv$ denotes the expected covariance matrix of $\yv$. To ensure the robustness of the inverse of $\Rv$, subarray averaging and diagonal loading techniques are employed.

The EBMV beamformer decomposes $\Rv$ through \YZ{eigenvalue} decomposition:
\begin{equation}
\Rv = \Ev\mathbf{\Lambda}\Ev^{\mathrm{H}},  
\end{equation}
where \(\Ev = [\ev_1, \ev_2, \dots, \ev_{{\it Ne}}]\) comprises the eigenvectors, and $\mathbf{\Lambda}$ is \YZ{the} diagonal matrix of eigenvalues. The beamforming weights in EBMV are computed by projecting $\wv_{\it MV}$ onto the signal subspace, which consists of the eigenvectors associated with the largest eigenvalues. These weights are given by:
\begin{equation}
    \wv_{\it EBMV}=\Ev_s \Ev_s{ }^{\mathrm{H}} \wv_{\it MV},
\end{equation}
where $\Ev_s=\left[\ev_1, \ev_2, \ldots, \ev_{{\it Num}}\right]$ with ${\it Num} < {\it Ne}$. 
The EBMV beamformed pixel \YZ{value} is then formulated as:
\begin{equation}
    \xv_{\it EBMV} = \wv^{\mathrm{H}}_{\it EBMV}\yv.
\end{equation}
Although speckle noise persists and image background completeness is often disrupted, EBMV effectively enhances spatial resolution.
For simplicity, the notation $\xv$ will be used throughout the remainder of this paper to represent the entire RF image reconstructed using EBMV.

\subsection{Denoising Diffusion Variance Imaging}
Diffusion models estimate clean images from noisy ones using both the measurement and the prior knowledge learned from training images~\cite{DDRM}. The measurement guides the sampling path, while the diffusion prior infers the information \YZ{affected} by the additive noise. Just as the diffusion sampling process can be modeled by a Stochastic Differential Equation (SDE)~\cite{song_solving_2022}, it involves numerous drift steps, so each sampling produces a different result. Although the generated estimations may appear quite similar under the same conditions, there will always be a certain degree of variation. 

Typically, the mean or the median of multiple samples, rather than a single sample, is used as the final result~\cite{chung2022scoreMRI} to enhance reliability, with the variance calculated to indicate the uncertainty of the sampling procedure. Interestingly, the variance map consistently shows a non-uniform pattern. In tasks where the ground truth lacks multiplicative noise, such as natural image restoration, the variance map tends to highlight the image boundaries~\cite{SNIPS}. \YZ{In contrast, in the presence of multiplicative noise, the variance map closely resembles an image with reduced noise}~\cite{EUSIPCO}.

\YZ{In US imaging, speckle, which is intrinsic to US images, can be modeled as multiplicative noise~\cite{CNNperdios2022,speckleModel2007Ng}:
\begin{equation}
    \ov = \mv \odot \pv,
    \label{Equ: x=mp}    
\end{equation}
where $\ov$ denotes the tissue reflectivity, $\mv$ represents the standard Gaussian multiplicative noise, and $\pv$ refers to the echogenicity map. Assuming that the EBMV-beamformed RF image is the sum of tissue reflectivity and additive noise,} the denoising model for diffusion sampling can be formulated as:
\begin{equation}
    \xv = \mv \odot \pv + \nv,
    \label{Equ: x=mp+n}    
\end{equation}
where $\xv$ represents the adaptively beamformed measurement, and $\nv \sim \mathcal{N}(\mathbf{0}, \gamma^2\Iv)$ denotes additive noise with a standard deviation of $\gamma$.

\YZ{Given the measurement $\xv$ and the noise level $\gamma$, denoising diffusion sampling can generate multiple estimations of $\ov$ (with the $c$-th sample denoted as $\hat\ov_c$).  We proposed an empirical model in \cite{EUSIPCO} that characterizes the output of a diffusion-based reconstruction as a function of the echogenicity $\pv$:
\begin{equation}
\hat\ov_c = \mv\odot\pv + \pv^{0.5} \odot\Gv_c.    
\end{equation}
In this model, $\Gv_c$ follows a standard normal distribution to account for stochasticity in the generative process. Relying on such a model, it can be easily verified that the variance of several diffusion-based US reconstructions provides information on the echogenicity.}
\section{Experimental validation on real data}\label{sec:Realdata}


\YZ{We focus on reconstructing US images from a single PW acquisition, using the results of coherent compounding from 75 DAS-beamformed PW images as the reference.} The \YZ{proposed} method was evaluated on the Plane Wave Imaging Challenge in Medical Ultrasound (PICMUS)~\cite{PICMUS} dataset. To demonstrate its performance, both qualitative and quantitative analyses were conducted using the \textit{in-vitro} \textit{Experimental Contrast (EC)} dataset. Additionally, we present qualitative results on the \textit{in-vivo} \textit{Carotid Cross-sectional (CC)} dataset.

EBMV beamforming was performed using the UltraSound ToolBox (USTB)~\cite{USTB}, with the following settings: \textit{active\_element\_criterium} (determining whether a transducer element is used) = 0.16, \textit{L\_element} (subarray size) = 80, \textit{K\_in\_lambda} (temporal averaging factor) = 0, \textit{refCoef} (diagonal loading constant) = 0.01, and \textit{gamma} (signal space criterion) = 0.05. All other parameters were kept at their default values. The EBMV beamformed RF signal was normalized to $[-1,1]$ to match the value range of the training set. 

\YZ{The employed diffusion model, initially pre-trained on ImageNet~\cite{ILSVRC15}, was fine-tuned on an in-house US dataset.
The fine-tuning dataset consists of 3551 (2,339 \emph{in vitro} + 1212 \emph{in vivo}) high-quality RF images with each reconstructed from 101 PWs using DAS. These images were collected using a TPAC Pioneer machine from the CIRS 040GSE phantom and a volunteer. 
Note that different from conventional autoencoder-based models for denoising~\cite{AutoEncoder}, diffusion-based models do not need paired data, only high-quality images are expected.}

\begin{figure}[tb]
    \centering
    \includegraphics[width=\linewidth]{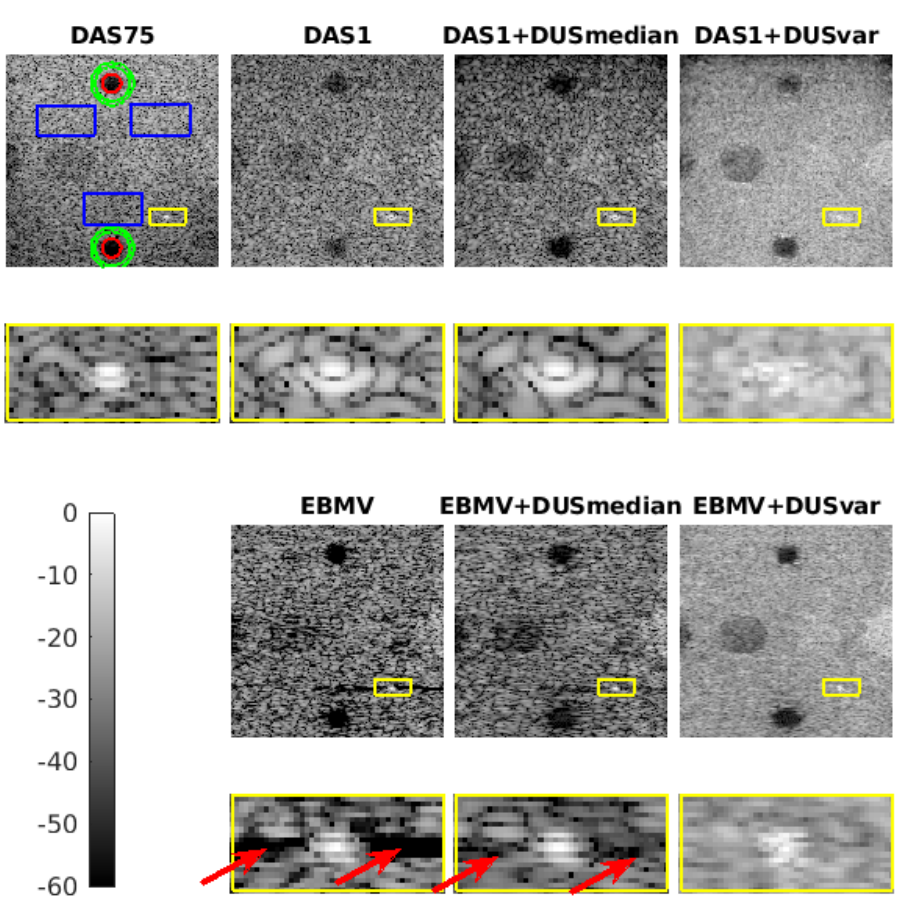}
    \caption{Comparison of reconstructed images on the PICMUS-\textit{EC} dataset. The colored boundaries outline the regions where the evaluation metrics are calculated.}
    \label{fig: EC_res}
\end{figure}

Each variance image was produced from 10 samples, and each sampling process (DUS) involved 50 iterations.  
The qualitative evaluation on the \textit{in-vitro} PICMUS-\textit{EC} dataset is shown in Fig.~\ref{fig: EC_res}. It consists of \YZ{seven} sets of images, each with a full view and a zoomed-in view to detail the scatterer region. The top-left is the reference reconstructed from 75 PWs using DAS, while the other images were reconstructed from a single PW. 
\begin{figure}[tb]
    \centering
    \includegraphics[width=\linewidth]{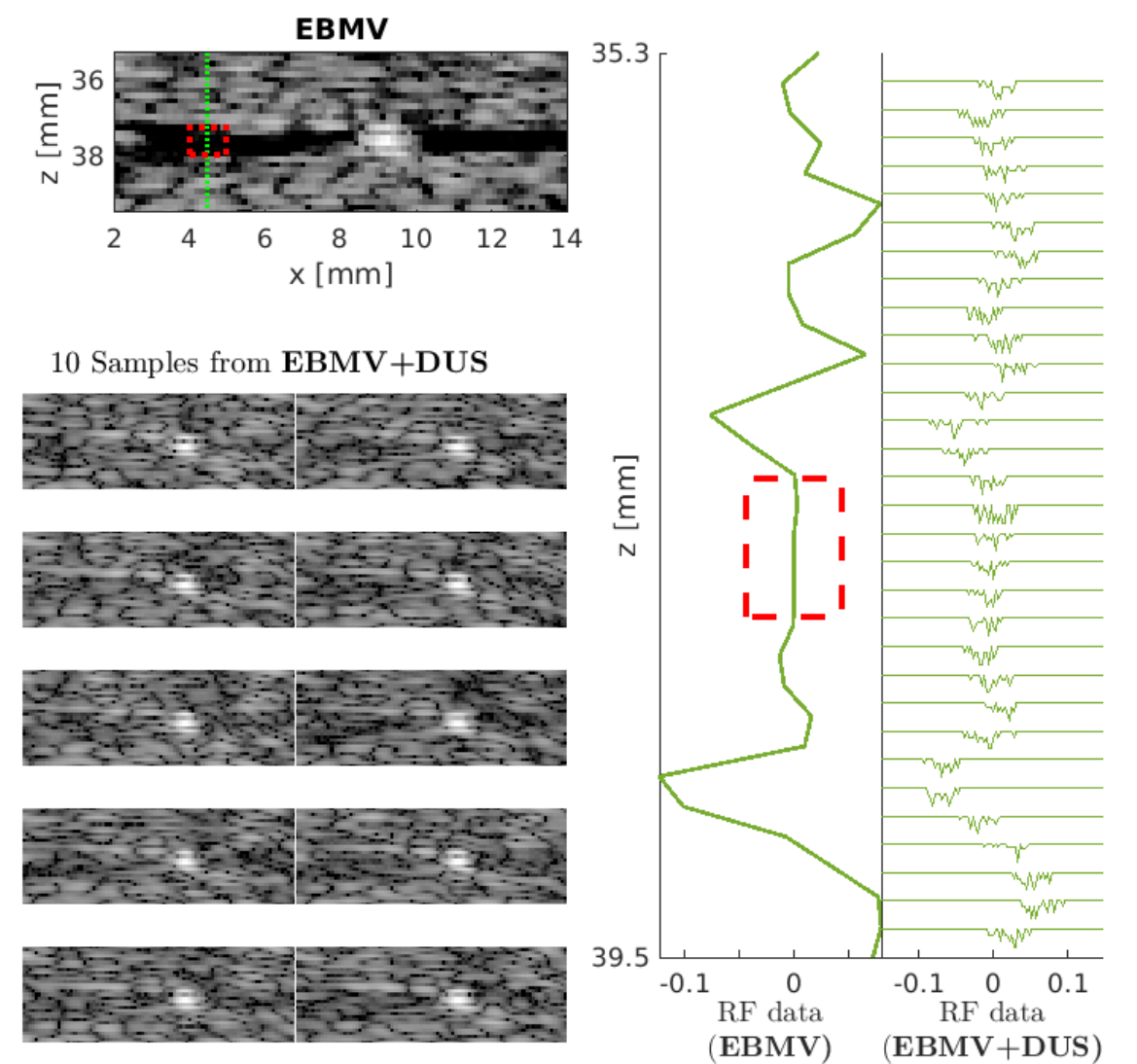}
    \caption{Statistical behavior of EBMV+DUS compared to EBMV on the PICMUS-\textit{EC} dataset. \YZ{The values at the position indicated by the dotted green line in the top-left image are compared in the right green plots. It shows that the variance of EBMV+DUS in the zero region enclosed by the dashed red box is non-zero. All grayscale images are in decibels with a dynamic range [-60,0].}}
    \label{fig: EC_ones}
\end{figure}
In the top row, the second image used DAS beamforming, and the following two images applied denoising diffusion to this DAS result: one shows the median of multiple samples, and the other the variance. In the bottom row, the first image was reconstructed with EBMV, and \YZ{it served as the measurement for denoising diffusion.} The second image displays the median result, whereas the fourth image, which utilizes the proposed EBMV+DUSvar method, illustrates the variance.

\YZ{A comparison of the anechoic regions and the zoomed-in scatterer views} reveals that EBMV+DUSvar significantly outperforms DAS1+DUSvar in \YZ{contrast and} spatial resolution. Additionally, while the completeness of the background is compromised in images reconstructed with EBMV or EBMV+DUSmedian (as indicated by the red arrows in Fig.~\ref{fig: EC_res}), the background is effectively restored in the image of EBMV+DUSvar.

The improvement in \YZ{contrast and} spatial resolution entirely relies on EBMV. The background recovery phenomenon is attributed to the generative stochasticity resulting from data mismatch. Specifically, the stochasticity in regions rendered anechoic by EBMV arises not from multiplicative noise but from discrepancies between the EBMV image and the training database. In the EBMV-beamformed image, the values in regions on either side of the scatterer are nearly zero, so denoising diffusion reconstruction for these regions relies entirely on the learned prior knowledge, as there are no measurements to guide the sampling process. As illustrated in Fig.~\ref{fig: EC_ones}, since the diffusion model was fine-tuned on high-quality US images with complete backgrounds, it tends to generate images with similarly complete backgrounds.

\begin{figure}[tb]
    \centering
    \includegraphics[width=\linewidth]{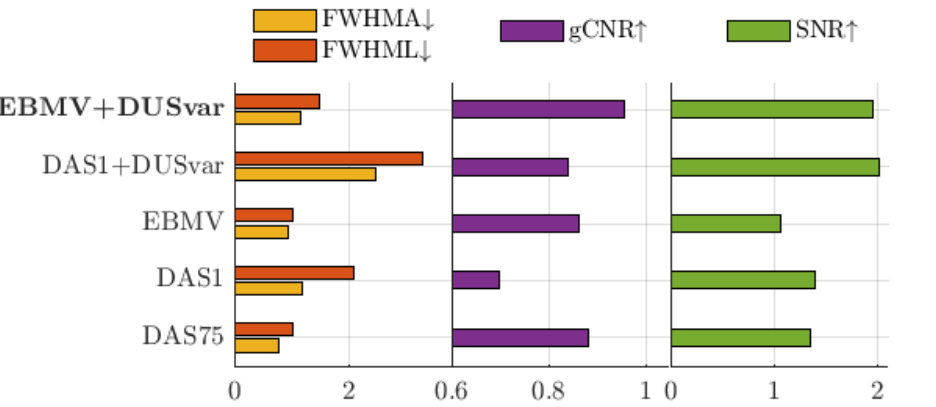}
    \caption{Quantitative comparison of the PICMUS-\textit{EC} dataset. A and L denote axial and lateral directions, respectively.}
    \label{fig: scores}
\end{figure}

Quantitative results evaluated in the colored regions on the DAS75 image are shown in Fig.~\ref{fig: scores}. Spatial resolution for the bright scatterer was assessed using the Full Width at Half Maximum (FWHM) in both axial and lateral directions. Contrast for the anechoic regions was evaluated using the generalized Contrast-to-Noise Ratio (gCNR)~\cite{gCNR}.
Background cleanliness was evaluated using the Signal-to-Noise Ratio (SNR).

As illustrated in Fig.~\ref{fig: scores}, the quantitative scores are consistent with the qualitative observations in Fig.~\ref{fig: EC_res}. This alignment confirms the significant improvement in spatial resolution and \YZ{contrast} of using EBMV compared to DAS1 for denoising diffusion variance imaging.

\begin{figure*}[tb]
    \centering
    \includegraphics[width=\linewidth]{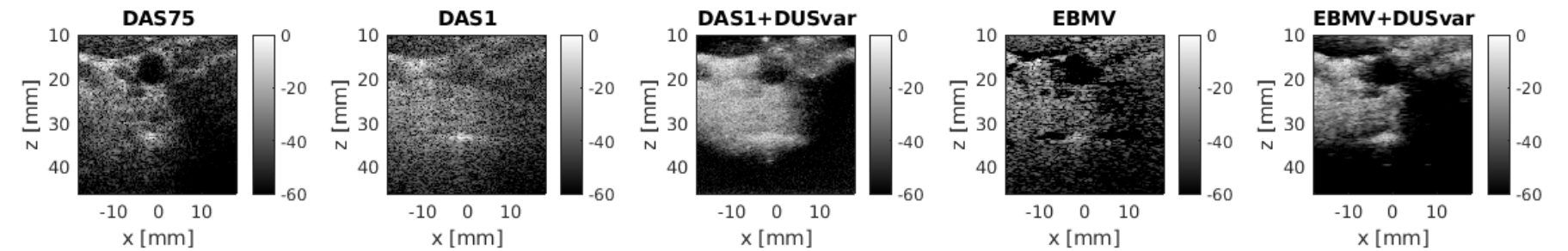}
    \caption{Comparison of reconstructed images on the PICMUS-\textit{CC} dataset.}
    \label{fig: Cross_res}
\end{figure*}

The qualitative evaluation on the \textit{in-vivo} PICMUS-\textit{CC} dataset, shown in Fig.~\ref{fig: Cross_res}, further confirms the feasibility of the proposed approach. Specifically, EBMV+DUSvar effectively smooths the image background compared to EBMV, while maintaining good spatial resolution and avoiding detail loss relative to DAS1+DUSvar.

\section{Conclusion}
This work proposes a model-based DL method combining EBMV beamforming with denoising diffusion variance imaging to address both electronic and speckle noise, enhancing the quality of single-PW US images.
Due to the high computational burden of solving inverse problems~\cite{EUSIPCO}, this work adopts a simpler denoising model to achieve faster sampling. To overcome the limited spatial resolution of DAS, EBMV beamforming is used instead. Note that while EBMV is utilized here, other adaptive beamformers could also be considered.
Experimental evaluations on the PICMUS-\textit{EC} and \textit{CC} datasets highlight the capacity of our method to balance spatial resolution, contrast, and background SNR. Furthermore, \YZ{our model} effectively recovers regions that are incorrectly rendered anechoic by EBMV, confirming its effectiveness.

\bibliographystyle{IEEEtran} 
\bibliography{IEEEabrv,references} 

\end{document}